\documentclass[journal]{IEEEtran} 
\usepackage{amsmath,amsfonts} 
\usepackage{algorithmic} 
\usepackage{algorithm} 
\usepackage{array}
\usepackage[caption=false,font=normalsize,labelfont=sf,textfont=sf]{ subfig} 
\usepackage{textcomp} 
\usepackage{stfloats} 
\usepackage{url}
\usepackage{verbatim} 
\usepackage{graphicx} 
\usepackage{tikz}
\usepackage{amsmath}
\usepackage{xcolor} 
\usepackage{hyperref}
\usepackage{booktabs}  % For better table aesthetics
\usepackage{caption}   % For customizing the table caption
\usepackage{array}     % For more flexible column formatting
\usepackage{soul}

\usepackage[utf8]{inputenc}
\usepackage{bibentry}
\nobibliography*

\definecolor{viridis1}{RGB}{117, 107, 177}
\definecolor{viridis2}{RGB}{126, 131, 186}
\definecolor{viridis3}{RGB}{124, 142, 196}
\definecolor{viridis4}{RGB}{108, 160, 200}
\definecolor{viridis5}{RGB}{91, 181, 201}
\definecolor{viridis6}{RGB}{100, 191, 186}
\definecolor{viridis7}{RGB}{155, 206, 130}
\definecolor{viridis8}{RGB}{218, 225, 93}
\definecolor{viridis9}{RGB}{250, 243, 117}

\begin{document}

    \title{A Staged Deep Learning Approach to Spatial Refinement in 3D Temporal Atmospheric Transport}

	\author{ \IEEEauthorblockN{\href{https://scholar.google.com/citations?user=ibl3c_gAAAAJ&hl=en}{M. Giselle Fern\'andez-Godino}\IEEEauthorrefmark{1}, Wai Tong Chung\IEEEauthorrefmark{2}, Akshay A. Gowardhan \IEEEauthorrefmark{1}, Matthias Ihme \IEEEauthorrefmark{2}, Qingkai Kong\IEEEauthorrefmark{1}, Donald D. Lucas\IEEEauthorrefmark{1}, and Stephen C. Myers\IEEEauthorrefmark{1} } \\ [15pt]
 
		\IEEEauthorblockA{\IEEEauthorrefmark{1}Lawrence Livermore National Laboratory, 7000 East Avenue, Livermore, CA 94550 \\ \{fernandez48, gowardhan1, lucas26, kong11, myers30\}@llnl.gov} \\ [5pt]

        \IEEEauthorblockA{\IEEEauthorrefmark{2}Stanford University, 440 Escondido Hall, Stanford, CA 94305 \\ \{mihme, wtchung\}@stanford.edu}

		\thanks{Manuscript received XX~Month~XXXX; revised XX~Month~XXXX. The co-authors are listed in alphabetical order. Details of the authors' contributions are provided in subsection Author Declarations (\ref{contributions}).}
	}

        % The paper headers
        \markboth{Artificial Intelligence in Geosciences,~Vol.~X, No.~X, November~XXXX}{Fern\'andez-Godino \MakeLowercase{\textit{et al.}}}

		\maketitle
		
		\begin{abstract} 
  
        High-resolution spatiotemporal simulations effectively capture the complexities of atmospheric plume dispersion in complex terrain. However, their high computational cost makes them impractical for applications requiring rapid responses or iterative processes, such as optimization, uncertainty quantification, or inverse modeling. To address this challenge, this work introduces the Dual-Stage Temporal Three-dimensional UNet Super-resolution (DST3D-UNet-SR) model, a highly efficient deep learning model for plume dispersion prediction. DST3D-UNet-SR is composed of two sequential modules: the temporal module (TM), which predicts the transient evolution of a plume in complex terrain from low-resolution temporal data, and the spatial refinement module (SRM), which subsequently enhances the spatial resolution of the TM predictions. We train DST3D-UNet-SR using a comprehensive dataset derived from high-resolution large eddy simulations (LES) of plume transport. We propose the DST3D-UNet-SR model to significantly accelerate LES simulations of three-dimensional plume dispersion by three orders of magnitude. Additionally, the model demonstrates the ability to dynamically adapt to evolving conditions through the incorporation of new observational data, substantially improving prediction accuracy in high-concentration regions near the source.

		\end{abstract}
		
		
		\section{Introduction}
        \IEEEPARstart{M}{odeling} atmospheric plume dispersion requires resolving the many scales of turbulence in fluid flows, a task for which large eddy simulations (LES) is particularly suited. LES captures the dynamics of turbulent eddies, yielding high-fidelity insights into atmospheric dispersion phenomena. However, the computational cost of LES is prohibitive for real-time applications, such as emergency response, and for iterative processes like optimization and uncertainty quantification (UQ) that demand large-scale simulations. To address these constraints, machine learning-based models have been developed, offering a trade-off between computational efficiency and predictive accuracy.
        
        In this context, we introduce the Dual-Stage Temporal 3D UNet Super-Resolution (DST3D-UNet-SR) model, a deep learning framework that significantly accelerates LES-based plume dispersion modeling while retaining high spatial resolution. DST3D-UNet-SR leverages a dual-stage neural network architecture: a temporal module (TM) and a spatial refinement module (SRM). The TM, built on a U-Net architecture augmented with ConvLSTM layers, processes temporal sequences of low-resolution input data to predict plume evolution over time while capturing dependencies across time steps. Following this, the SRM enhances the spatial resolution of the TM outputs using a 3D U-Net architecture optimized with skip connections and transposed convolutions, enabling a fourfold increase in resolution. By combining these two specialized modules, DST3D-UNet-SR significantly reduces computational costs by orders of magnitude while retaining the fidelity of turbulence-resolving LES simulations.
        
		\section{Related Work} The field of atmospheric plume dispersion modeling has seen significant advancements through the integration of machine learning techniques. While traditional approaches such as LES provide unparalleled accuracy, their computational demands have motivated the development of alternative methods that prioritize efficiency without sacrificing reliability. In this section, we review prior work on integrating convolutional neural networks (CNNs) and Long Short-Term Memory (LSTMs) for spatiotemporal modeling in atmospheric sciences.

        CNNs have shown remarkable success in spatial pattern prediction tasks, outperforming traditional methods in various domains, including atmospheric modeling \cite{camps2021deep}. In our prior work \cite{fernandez2023predicting}, we demonstrated the applicability of CNNs to spatial deposition modeling, highlighting their ability to reconstruct fine-scale features. Additionally, in a related study \cite{kong2021deep}, we explored the use of CNNs as feature extractors for seismic waveform data, demonstrating the effectiveness of CNN in enhancing encoded features for tasks such as event discrimination and phase picking. Of particular relevance is the Super-Resolution Convolutional Neural Network (SRCNN) developed by Dong et al. (2015) \cite{dong2015image}, a foundational DL framework for enhancing the spatial resolution of low-quality images. SRCNN directly learns an end-to-end mapping from low-resolution to high-resolution representations, leveraging its deep CNN architecture to achieve superior single-image super-resolution. We studied this technique in previous work \cite{chung2023turbulence}, establishing that super-resolution techniques perform well in turbulent flows up to 16x-refinement, indicating that our 4x-refinement dual-stage model is well within this effective range.

        Long Short-Term Memory (LSTM) have proven to be highly effective in modeling temporal sequences involving atmospheric plume dispersion. LSTMs effectively capture long-term dependencies in time-series data, crucial for forecasting and assessing plume dispersion dynamics. The study by Marcos et al. (2024) \cite{marcos2024exploring} highlights the great performance achieved by LSTMs in modeling the intricate temporal evolution of plume dispersion dynamics, specifically demonstrating their application in predicting the atmospheric dispersion of radioactive plumes during nuclear power plant emergencies. Similarly, our prior research \cite{fernandez2021accelerating, garcia2022uncertainty, wang2021stressnet, wang2024spatiotemporal} has shown that LSTMs achieve high accuracy in forecasting temporal patterns, emphasizing their potential to enhance real-time decision-making scenarios.

        Previous research has explored the integration of CNNs and LSTMs for spatiotemporal modeling; however, the DST3D-UNet-SR model introduces significant advancements. The model employs LSTMs for temporal predictions using low-resolution input data, effectively and efficiently capturing temporal dependencies through the TM before applying the SRM. DST3D-UNet-SR model integrates a 3D U-Net architecture within its SRM to achieve a fourfold enhancement in spatial resolution. A defining feature of DST3D-UNet-SR is its modular design, which allows for the choice to apply super-resolution sequentially at each timestep or collectively after completing the temporal modeling. This modularity ensures that enhancements to either the TM or SRM can be independently achieved, streamlining model refinement and adaptation for future applications. DST3D-UNet-SR unifies temporal modeling and spatial refinement in a single framework, offering a more comprehensive solution compared to prior models, which often address isolated aspects or lack super-resolution capabilities.
        
        For example, Tsokov et al. (2022) \cite{tsokov2022hybrid} presented a hybrid model integrating CNNs and LSTMs for air pollution forecasting in two dimensions. Their Deep-Air model captures spatial and temporal dependencies within a single framework and is conceptually similar to the baseline HRTM model used in this study (see section ~\ref{results} instead of the DST3D-UNet-SR staged approach. Moreover, the SRM incorporates ConvLSTM layers to further refine spatial features, a capability not present in Deep-Air.
        
        Similarly, Ding et al. (2021) introduced a hybrid CNN-LSTM model to predict daily PM2.5 concentrations by leveraging spatiotemporal correlations between Beijing and its neighboring cities. While effective for analyzing 2D spatial data, their approach does not incorporate super-resolution techniques or U-Net architectures, which limits its applicability to high-resolution 3D atmospheric modeling.

        To the best of our knowledge, this work is the first to integrate U-Net architectures, super-resolution techniques, LSTMs, and CNNs into a modular framework, the DST3D-UNet-SR, specifically designed for modeling the 3D temporal evolution of plume dispersion over complex terrain.

        \section{Data} \label{Data}

        The simulations, referred in this work as \textit{ground truth}, were originally designed to replicate similar conditions of the REACT (RElease ACTivity) experiment, conducted in October 2022 at the Nevada National Security Site (NNSS), where real-time xenon sensors monitored radiotracer releases \cite{wharton2023capturing, stave2024real} (Fig. \ref{sensor_location}). Xenon is a reliable indicator of nuclear explosions, particularly underground nuclear tests. We collaborate with researchers who utilize data from the International Monitoring System or IMS, a network designed to detect and verify nuclear explosions globally to ensure compliance with the Comprehensive Nuclear-Test-Ban Treaty or CTBT.
        
\begin{figure}[ht!] \centering \noindent\includegraphics[width=0.45\textwidth]{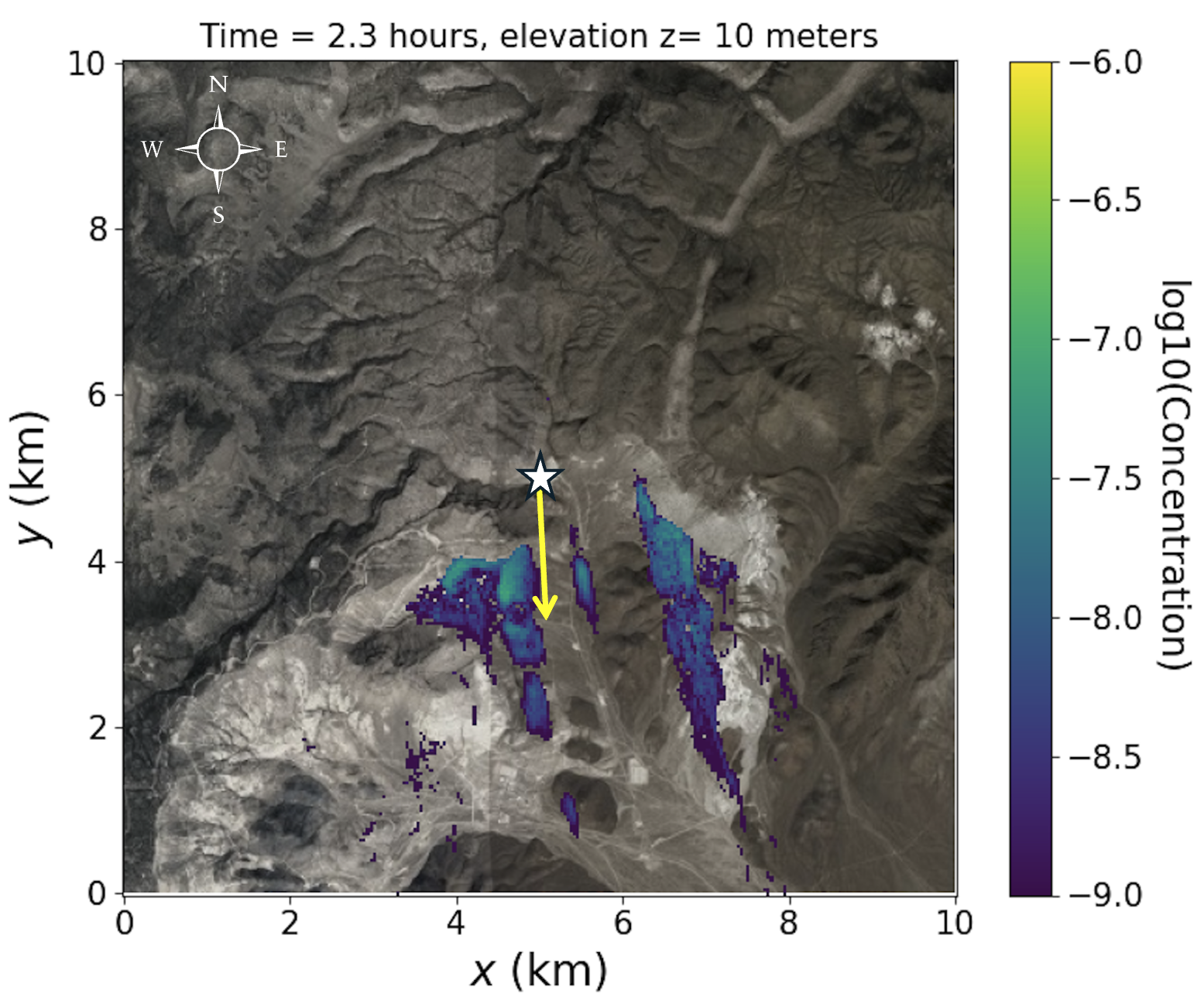} \caption{An example of a xenon field at 10 m slice above the ground level with $w_s$ = 7.8 m/s, $w_d$ = 357$^\circ$ after 2.3 hours following the start of the release. The white star marks the release location, while the yellow arrow indicates the wind direction. The complex terrain, characterized by mountains and valleys, is shown in gray scale.} \label{disc} \end{figure}

        A continuous release of Xe-133 was simulated, accounting for its half-life of 5.2 days. However, given that the simulation spans only five hours, the decay rate is negligible and does not impact the results within the considered timeframe.
        The xenon release occurs at the center of the complex terrain domain of $10$~km $\times$ $10$~km, i.e. $(x_{\text{release}},y_{\text{release}}) = (5~ \text{km}, 5~ \text{km})$, as shown in Figure \ref{disc} using a white star symbol. The gas is released continuously from a point source with a unit mass that can be scaled to cover specified source amounts. Xenon is used as a tracer in release experiments because it is a non-reactive, inert noble gas with a non-toxic nature that enables precise sensor measurements as later shown in section \ref{validation}. 
        
        Aeolus~\cite{gowardhan_large_2021} dispersion code was used to perform the simulations. Aeolus is an advanced LES model specifically engineered to predict flow fields and the transport of gases and particles in areas characterized by urban environments and complex terrain, which poses unique challenges for atmospheric modeling. By employing the LES methodology, the model effectively solves the three-dimensional incompressible Navier-Stokes equations, allowing for a detailed representation of turbulent flow dynamics. The transport of tracers is intricately modeled using an in-line coupled Lagrangian dispersion approach, where each marker particles symbolizes a defined substance, derived from the source term applicable to both gaseous emissions and particulate matter. These marker particles are dynamically transported by the wind and turbulence calculated by the model, reflecting real-world atmospheric conditions. As they traverse the landscape, these particles interact with the surrounding terrain, leading to mass loss through processes such as gravitational settling, where heavier particles fall to the ground, and various non-gravitational deposition mechanisms applicable to gases, including adsorption and impaction, which further influence the distribution and concentration of gas and particles in the environment. This comprehensive modeling framework enables Aeolus to provide critical insights into tracer dispersion, aiding in environmental assessments and the development of effective mitigation strategies.
 
        The dataset consists of 100 runs: 80 for training, 10 for validation, and 10 for testing. The simulations varied the parameters wind speed ($w_s$) and wind direction ($w_d$) selected using Latin Hypercube Sampling (LHS)~\cite{stein1987large}, with $w_s$ ranging from 1.5 to 10 m/s and $w_d$ from 340$^\circ$ to 360$^\circ$, representing the predominant wind conditions at this site. The complex terrain, characterized by mountains and valleys, presents a challenging problem. The ML model must learn the discontinuities in flow concentration. Figure~\ref{disc} presents a 2D slice of one of the concentration fields used for training. 

        The dataset was curated to ensure quality and relevance for training the ML model. From the 330 time steps extracted from Aeolus, each one minute apart, we retained 33 time steps for training at 10-minute intervals, resulting in a training duration of 5.5 hours. The simulations were run using a resolution of $(z,y,x)=200 \times 250 \times 250$ cells, each cell representing $20 \, \text{m} \times 40 \, \text{m} \times 40 \, \text{m}$. Concentrations are recorded at the cell center. A height of $z = 0$ corresponds to ground level and is unrelated to sea level. Plume concentration data is presented logarithmically to facilitate visualization.
        
        Initially, the simulation domain comprised a volume of $[z, x, y] = [4, 10, 10]$ km, which was then cropped to $[2, 5, 5]$ km for training purposes due to the plume's predominant movement towards the southeast quadrant and altitudes less than 2 km, influenced by wind velocity. At this stage, the dataset is resized into two distinct resolutions using the \textit{resize} function from the Python package \textit{skimage.transform}. The low-resolution data is formatted with dimensions $\left[\text{number of time steps}, \text{height}, \text{width}, \text{depth}\right] = \left[33, 8, 32, 32\right]$, where the height, width, and depth correspond to the $z$, $x$, and $y$ spatial dimensions, respectively. Each grid cell represents $250 , \text{m} \times 156.3 , \text{m} \times 156.3 , \text{m}$. The high-resolution data, on the other hand, is stored with dimensions $\left[\text{number of time steps}, \text{height}, \text{width}, \text{depth}\right] = \left[33, 32, 128, 128\right]$, where the $z$, $x$, and $y$ spatial dimensions correspond to grid cells of $62.5 , \text{m} \times 39.1 , \text{m} \times 39.1 \text{m}$. Both datasets are stored in 32-bit format, with the low-resolution dataset totaling 0.1 GB and the high-resolution dataset totaling 6.6 GB.
        
        The Aeolus LES model, as a data source, provides us with sufficient data cubes to train our model effectively. However, for cases with sparse data, data symmetries can be employed for data augmentation \cite{fernandez2019use} and performance enhancement. To predict gas dispersion, data can often be augmented for different wind directions by rotating the plume. However, in our case, this approach is not applicable due to asymmetries in the terrain.

		\section{DST3D-UNet-SR Architecture} \label{Arch}
		
\begin{figure}[ht]
            \centering
            \begin{tikzpicture}[
                conv/.style={draw, rectangle, minimum width=3cm, minimum height=1cm, text centered, text width=3cm, node distance=1.5cm, fill=viridis4},
                convtrans/.style={draw, rectangle, minimum width=3cm, minimum height=1cm, text centered, text width=3cm, node distance=1.5cm, fill=viridis6},
                pool/.style={draw, rectangle, minimum width=3cm, minimum height=1cm, text centered, text width=3cm, node distance=1.5cm, fill=viridis9},
                skip/.style={draw, rectangle, minimum width=3cm, minimum height=1cm, text centered, text width=3cm, node distance=1.5cm, fill=viridis3},
                tm_convlstm/.style={draw, rectangle, minimum width=3cm, minimum height=1cm,text centered, text width=3cm, node distance=1.5cm, fill=viridis3},
                ]
                
                % Temporal Module
                \node (tm_title) at (0, 8) {\textbf{Temporal Module}};
                
                \node (tm_enc1) [conv, below of=tm_title, yshift=0cm] {Conv3D: (5, 7$\times$16)};
                \node (tm_pool1) [pool, below of=tm_enc1] {MaxPool3D};
                
                \node (tm_enc2) [conv, below of=tm_pool1] {Conv3D: (7$\times$16, 7$\times$32)};
                \node (tm_pool2) [pool, below of=tm_enc2] {MaxPool3D};
                
                \node (tm_enc3) [conv, below of=tm_pool2] {Conv3D: (7$\times$32, 7$\times$64)};
                \node (tm_pool3) [pool, below of=tm_enc3] {MaxPool3D};
                
                \node (tm_convlstm) [tm_convlstm, below of=tm_pool3] {ConvLSTM: (7$\times$64, 7$\times$64)};
                
                \node (tm_dec1) [convtrans, below of=tm_convlstm] {ConvTranspose3D: (7$\times$64, 7$\times$32)};
                \node (tm_dec2) [convtrans, below of=tm_dec1] {ConvTranspose3D: (7$\times$32, 7$\times$16)};
                \node (tm_dec3) [convtrans, below of=tm_dec2] {ConvTranspose3D: (7$\times$16, 1)};
                
                % Arrows for Temporal Module
                \draw[->] (tm_enc1) -- (tm_pool1);
                \draw[->] (tm_pool1) -- (tm_enc2);
                \draw[->] (tm_enc2) -- (tm_pool2);
                \draw[->] (tm_pool2) -- (tm_enc3);
                \draw[->] (tm_enc3) -- (tm_pool3);
                \draw[->] (tm_pool3) -- (tm_convlstm);
                \draw[->] (tm_convlstm) -- (tm_dec1);
                \draw[->] (tm_dec1) -- (tm_dec2);
                \draw[->] (tm_dec2) -- (tm_dec3);
                
                % Spatial Refinement Module
                \node (srm_title) at (4, 8) {\textbf{Spatial Refinement Module}};
                
                \node (srm_enc1) [conv, below of=srm_title, yshift=0cm] {Conv3D: (1, 7$\times$16)};
                \node (srm_pool1) [pool, below of=srm_enc1] {MaxPool3D};
                
                \node (srm_enc2) [conv, below of=srm_pool1] {Conv3D: (7$\times$16, 7$\times$32)};
                \node (srm_pool2) [pool, below of=srm_enc2] {MaxPool3D};
                
                \node (srm_adjust) [skip, below of=srm_pool2] {Skip Connections: (7$\times$16, 7$\times$32)};
                
                \node (srm_dec1) [convtrans, below of=srm_adjust] {ConvTranspose3D: (7$\times$32, 7$\times$32)};
                \node (srm_dec2) [ convtrans, below of=srm_dec1] {ConvTranspose3D: (7$\times$32, 7$\times$16)};
                \node (srm_dec3) [convtrans, below of=srm_dec2] {ConvTranspose3D: (7$\times$16, 7$\times$8)};
                \node (srm_dec4) [convtrans, below of=srm_dec3] {ConvTranspose3D: (7$\times$8, 1)};
                
                % Arrows for Spatial Refinement Module
                \draw[->] (srm_enc1) -- (srm_pool1);
                \draw[->] (srm_pool1) -- (srm_enc2);
                \draw[->] (srm_enc2) -- (srm_pool2);
                \draw[->] (srm_pool2) -- (srm_adjust);
                \draw[->] (srm_adjust) -- (srm_dec1);
                \draw[->] (srm_dec1) -- (srm_dec2);
                \draw[->] (srm_dec2) -- (srm_dec3);
                \draw[->] (srm_dec3) -- (srm_dec4);
                
            \end{tikzpicture}
            \caption{DualStage Temporal 3D UNet-SR Model Architecture}
            \label{fig:dualstage_architecture}
        \end{figure}
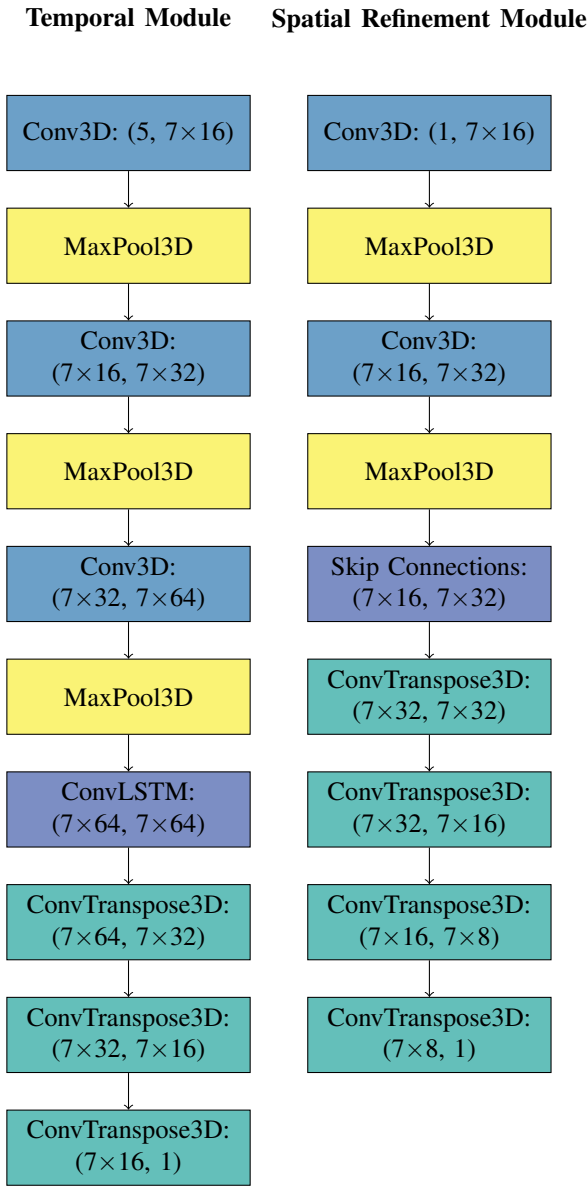

        The DST3D-UNet-SR model is designed to predict the temporal and spatial resolution of 3D plume dispersion by integrating two specialized neural network architectures: a temporal module (TM) for predicting the transient evolution from low-resolution data and spatial refinement module (SRM) to enhance the spatial resolution. The separation into these two modules provides several advantages, including flexibility in design and training, modular upgrades, and targeted improvements. This modular approach allows each component to be independently optimized and updated, ensuring that improvements in one module do not impact the other. Therefore, it facilitates the incorporation of new techniques and advancements in either temporal prediction or spatial refinement without requiring a complete redesign of the entire ML model. This section details the architectures and training strategies of the TM and the SRM. Figure \ref{fig:dualstage_architecture} is a schematic of each module's architecture.

		\subsection{Temporal Module and Training Strategy} 
		
        The temporal module (TM) in our DST3D-UNet-SR model is designed to capture and predict the evolution of atmospheric dispersion phenomena in 3D. This model employs a U-Net architecture for spatial features enhanced with ConvLSTM layers, enabling it to handle the spatiotemporal dependencies inherent in the data effectively. The TM inputs are sequences shaped as $(\text{batch size}, 5, 8, 32, 32)$, representing a temporal window of five time steps, each with a spatial resolution of $8 \times 32 \times 32$. The encoder section of the model progressively reduces the spatial dimensions while increasing the number of channels, using three convolutional blocks followed by max-pooling and 20\% dropout layers to prevent overfitting. The ConvLSTM layer at the bottleneck captures the temporal dependencies across the encoded representations. The decoder section then reconstructs the high-level features back to the original spatial dimensions, incorporating skip connections from the encoder to preserve spatial details. This process results in an output shaped as $(\text{batch size}, 1, 8, 32, 32)$ representing a single time step, specifically the one that follows the five previous time steps provided as input, allowing the model to predict the temporal evolution of the plume concentration autoregressively. Further details on the TM architecture are provided in Appendix \ref{TM}. 

        The training strategy for the TM employs a sliding-window approach across the temporal dimension. Each training sequence consists of five consecutive time steps used to predict the next time step in the sequence. This ensures the model learns to capture temporal dependencies effectively. Various configurations were tested, and a window size of five time steps was found to capture sufficient temporal context while maintaining computational efficiency and model performance. The training process uses the mean squared error (MSE) loss function, and the parameters are updated using the Adam optimizer \cite{kingma2014adam}, with a learning rate scheduler to adapt as validation loss plateaus. The TM model was trained independently of the SRM, on an Apple MacBook Pro equipped with an M2 Max chip and 32GB of memory, utilizing the system's 12-core CPU, with a training time of approximately 1.2 minutes and 1000 epochs.

        During inference, the TM operates in an autoregressive prediction framework. Initially, the model receives a five-time-step input of shape $(1, 5, 8, 32, 32)$ consisting entirely of ground truth values to predict the subsequent time step, $t+1$ of shape $(1, 1, 8, 32, 32)$. This prediction is then incorporated into the input sequence for the next iteration, replacing the oldest time step while retaining the remaining ground truth values. As predictions progress, the reliance on ground truth data diminishes, with subsequent inputs increasingly comprising model-predicted values. By $t+5$, the model operates in a fully predictive regime, where all input time steps are derived exclusively from prior predictions. 
        
		\subsection{Spatial Refinement Module and Training Strategy}
		
		The spatial refinement module (SRM) in the DST3D-UNet-SR model focuses on enhancing the spatial resolution of under-resolved 3D atmospheric simulation data through a super-resolution approach \cite{dong2015image}. The module employs a U-Net architecture specifically designed for 3D data, enabling it to upscale coarse spatial resolutions to high-resolution outputs. The SRM processes input tensors of shape $(\text{batch size}, 8, 32, 32)$, representing a low-resolution 3D spatial grid, and produces high-resolution outputs of shape $(\text{batch size}, 32, 128, 128)$. Its encoder consists of convolutional layers that progressively reduce spatial dimensions while increasing feature channels. Each convolutional block is followed by batch normalization and LeakyReLU activation functions, with max-pooling layers for spatial downsampling. Skip connections between encoder and decoder layers ensure that spatial information is retained during the reconstruction process. The decoder employs transposed convolutional layers to upscale feature maps, achieving a fourfold enhancement in spatial resolution. While higher upsampling factors are feasible, they can exceed the model's capacity to capture detailed features effectively, as discussed in \cite{chung2023turbulence}. Further architectural details of the SRM are provided in Appendix \ref{SRM}.

        The SRM is trained using supervised learning, with low-resolution data generated from high-resolution ground truth through average pooling. During training, the model receives low-resolution inputs and predicts high-resolution outputs, which are optimized to match the ground truth using the mean squared error (MSE) loss function. The Adam optimizer \cite{kingma2014adam} is used for parameter updates, with a learning rate scheduler applied to reduce the rate when the validation loss plateaus. This training approach ensures convergence while capturing fine spatial details and preventing overfitting. The SRM was trained independently of the TM on LLNL’s Lassen high-performance computing system, utilizing 4 NVIDIA Volta GPUs in parallel. The training process was completed in approximately 3 hours over 100 epochs.
        
		\subsection{Integration of Temporal and Spatial Refinement Modules} 
		
		The combined workflow of the TM and SRM can be summarized as follows: 

        \begin{enumerate} 
        
        \item The TM uses an autoregressive process to make 10-minute increment predictions. It receives a batch of low-resolution sequences, each with the shape $(\text{batch size}, 5, 8, 32, 32)$. These sequences represent a sliding window of five consecutive time steps at 10-minute intervals.
        
        \item The TM processes these sequences and outputs a batch of predicted low-resolution frames, each of shape $(\text{batch size}, 1, 8, 32, 32)$. This single frame represents the prediction for the next time step following the five previous time steps provided as input.
        
        \item The next TM's prediction input consists of the last four time steps of the previous input concatenated with the latest TM's prediction. In other words, the next time step input is constructed by removing the first time step from the previous input and appending the prediction of the latest time step.
        
        \item For the first five predictions, the TM includes low-resolution ground truth data in its input sequences. After the initial five predictions, all subsequent predictions are based solely on the model's previous predictions. 
        
        \item The TM's inputs and the predicted low-resolution frames from the fifth element of each sequence are then fed into the SRM. 
        
        \item The SRM processes these predicted frames and outputs high-resolution frames, each of shape $(\text{batch size}, 32, 128, 128)$. 
        \end{enumerate} 

\begin{figure*}[ht!] 
        \centering \noindent\includegraphics[width=0.8\textwidth]{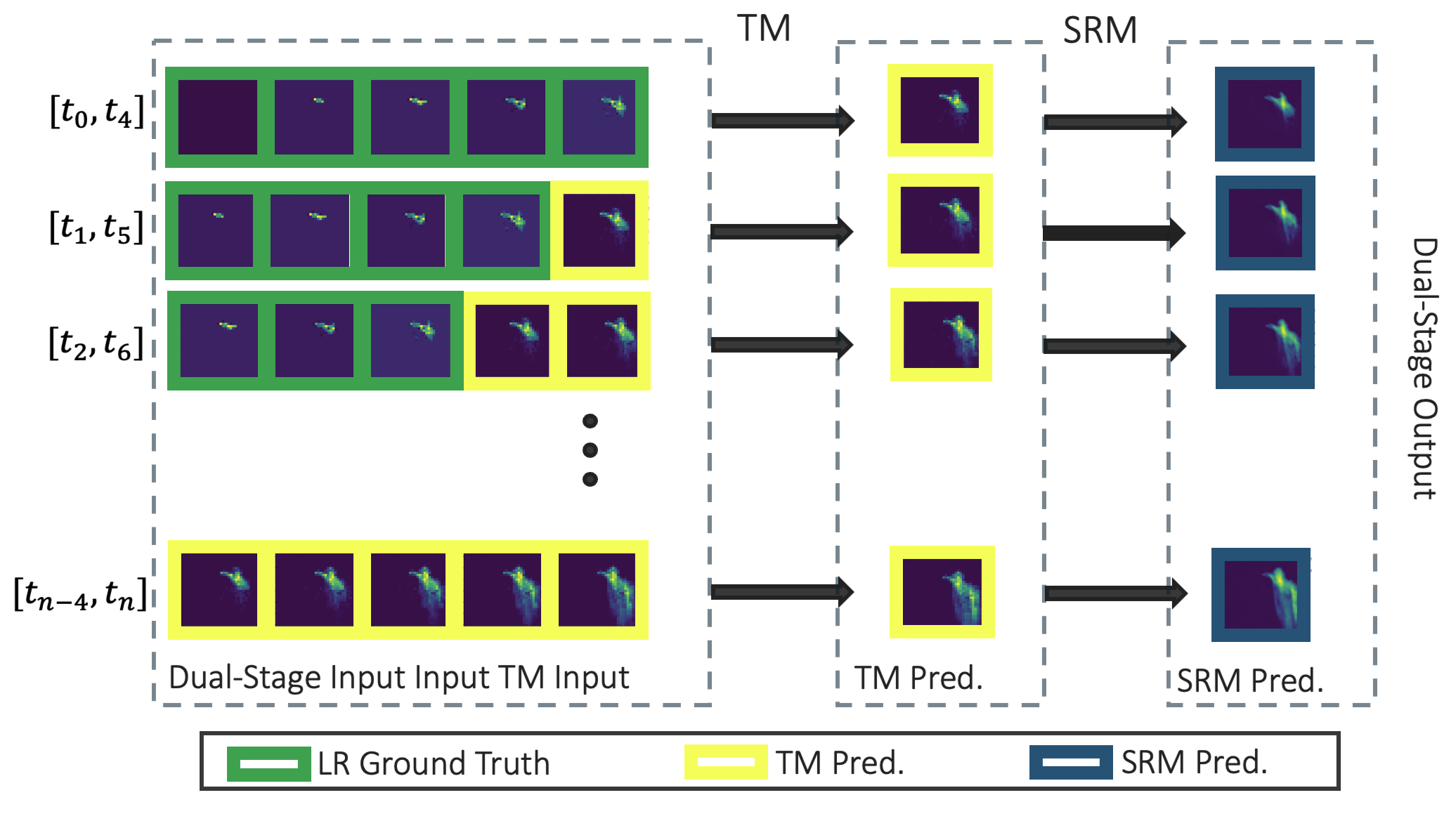} \caption{DST3D-UNet-SR workflow as combination of the TM and SRM. The TM receives a batch of low-resolution sequences and outputs a batch of predicted low-resolution frames. The next TM's prediction input consists of the last four time steps of the previous input concatenated with the latest TM's prediction. The TM's inputs and the predicted low-resolution frames from the fifth element of each sequence are then fed into the SRM, which outputs the corresponding high-resolution frames.} \label{pipeline} 
        \end{figure*}

        The workflow shown above is also depicted in Figure \ref{pipeline}.
        
		\section{Results} \label{results}

        In this section, we assess the DST3D-UNet-SR model’s effectiveness in predicting plume dispersion. The evaluation includes comparisons of the predicted concentrations against ground truth values in the \textit{x-}, \textit{y-}, and \textit{z-} directions. To evaluate the performance of the DST3D-UNet-SR, we compare it against the high-resolution temporal model (HRTM), which shares the same architecture as the TM but operates directly on high-resolution volumetric data. The HRTM serves as a baseline for comparison and was trained with a parameter count comparable to that of the DST3D-UNet-SR model, approximately 4 million. Notably, the HRTM requires high-resolution input data, making it more computationally expensive, although more accurate for the initial time steps. In contrast, the DST3D-UNet-SR model takes low-resolution input data, which it enhances to high-resolution through its integrated SRM. The HRTM was trained in parallel using four NVIDIA V100 GPUs (each with 16GB of memory) on LLNL's Lassen high-performance computing system, with a total training time of approximately 7 hours over 100 epochs. Detailed information about the HRTM can be found in Appendix~\ref{B}.
  
\begin{figure}[ht!] 
        \centering \noindent\includegraphics[width=0.5\textwidth]{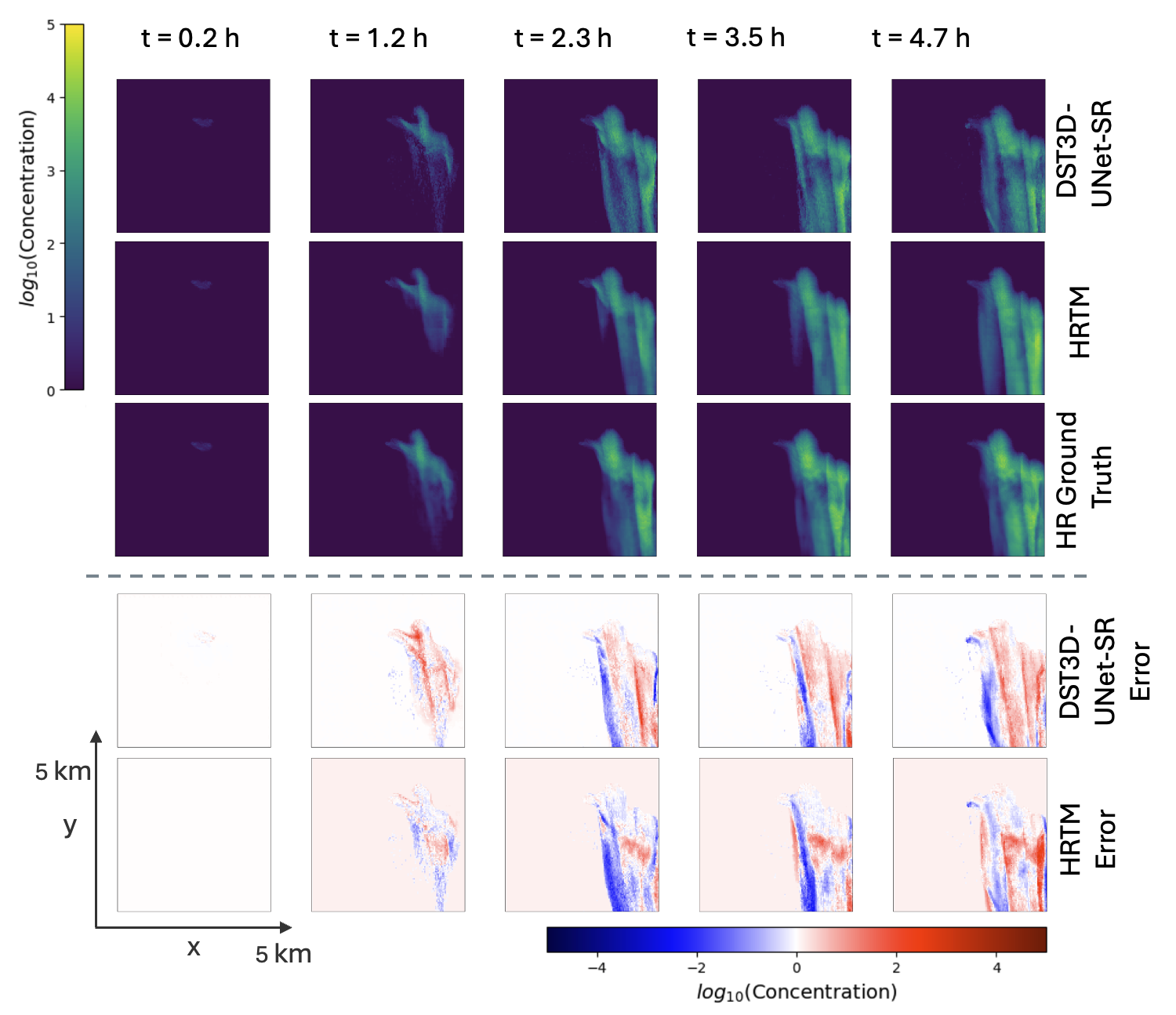} \caption{Averaged concentration in the \textit{x-y} plane at various time steps, comparing DST3D-UNet-SR and HRTM predictions (top two rows) with ground truth (third row). The errors are also shown in the figure (two bottom rows). The DST3D-UNet-SR error is calculated as the difference between DST3D-UNet-SR predictions and ground truth. Similarly, the HRTM error is calculated as the difference between HRTM predictions and ground truth. Variations are depicted across the \textit{x-}direction and \textit{y-}direction within ranges of 0 to 5 km.} \label{z_mean} 
        \end{figure}

        Figures \ref{x_mean}, \ref{y_mean}, and \ref{z_mean} show the averaged concentration in the \textit{x-}, \textit{y-}, and \textit{z-}planes, respectively, at various time steps for a single test case with $w_s$ = 5.7 m/s and $w_d$ = 350.5$^{\circ}$. These figures compare DST3D-UNet-SR and HRTM predictions (two top rows) with ground truth (third row). Variations are depicted across the other two dimensions within specified ranges, with appropriate spatial resolutions. Specifically, Figure \ref{x_mean} shows variations across the \textit{y-} and \textit{z-}plane within ranges of 0 to 5 km and 0 to 2 km, respectively, with spatial resolutions of 0.04 km for \textit{y} and 0.02 km for \textit{z} vertical slices. Figure \ref{y_mean} shows variations across the \textit{x-} and \textit{z-}-plane within the same ranges and resolutions. Figure \ref{z_mean} shows variations across the \textit{x-} and \textit{y-}plane within a range of 0 to 5 km, with a spatial resolution of 0.04 km for both dimensions.

        \begin{figure*}[ht!] 
        \centering \noindent\includegraphics[width=\textwidth]{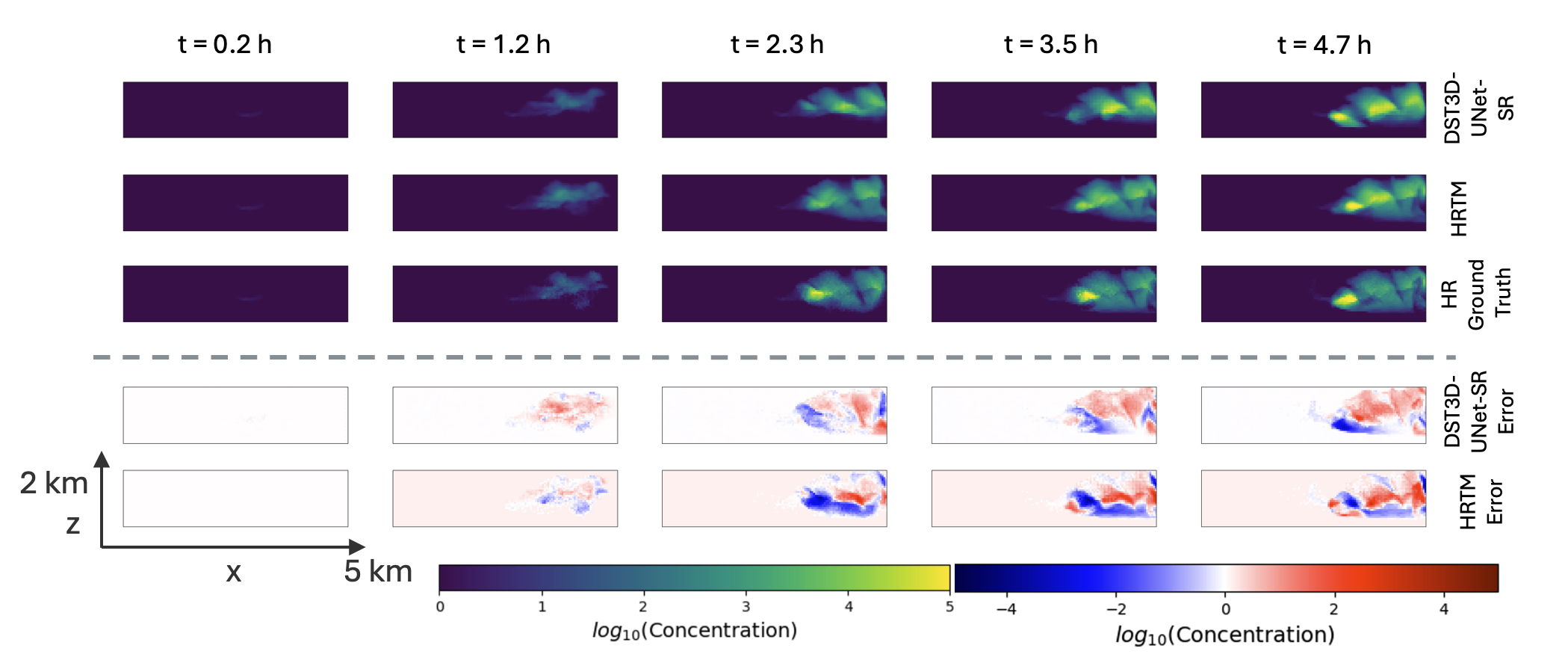} \caption{Averaged concentration in the \textit{x-z} plane at various time steps, comparing DST3D-UNet-SR and HRTM predictions (two top rows) with ground truth (third row). The errors are also shown in the figure (two bottom rows). The DST3D-UNet-SR error is calculated as the difference between DST3D-UNet-SR predictions and ground truth. Similarly, the HRTM error is calculated as the difference between HRTM predictions and ground truth. Variations are shown across the \textit{x} and \textit{z} dimensions within a range of 0 to 5 km and the \textit{z-}dimension ranging from 0 to 2 km.} \label{x_mean} 
        \end{figure*} 
        
        \begin{figure*}[ht!] 
        \centering \noindent\includegraphics[width=\textwidth]{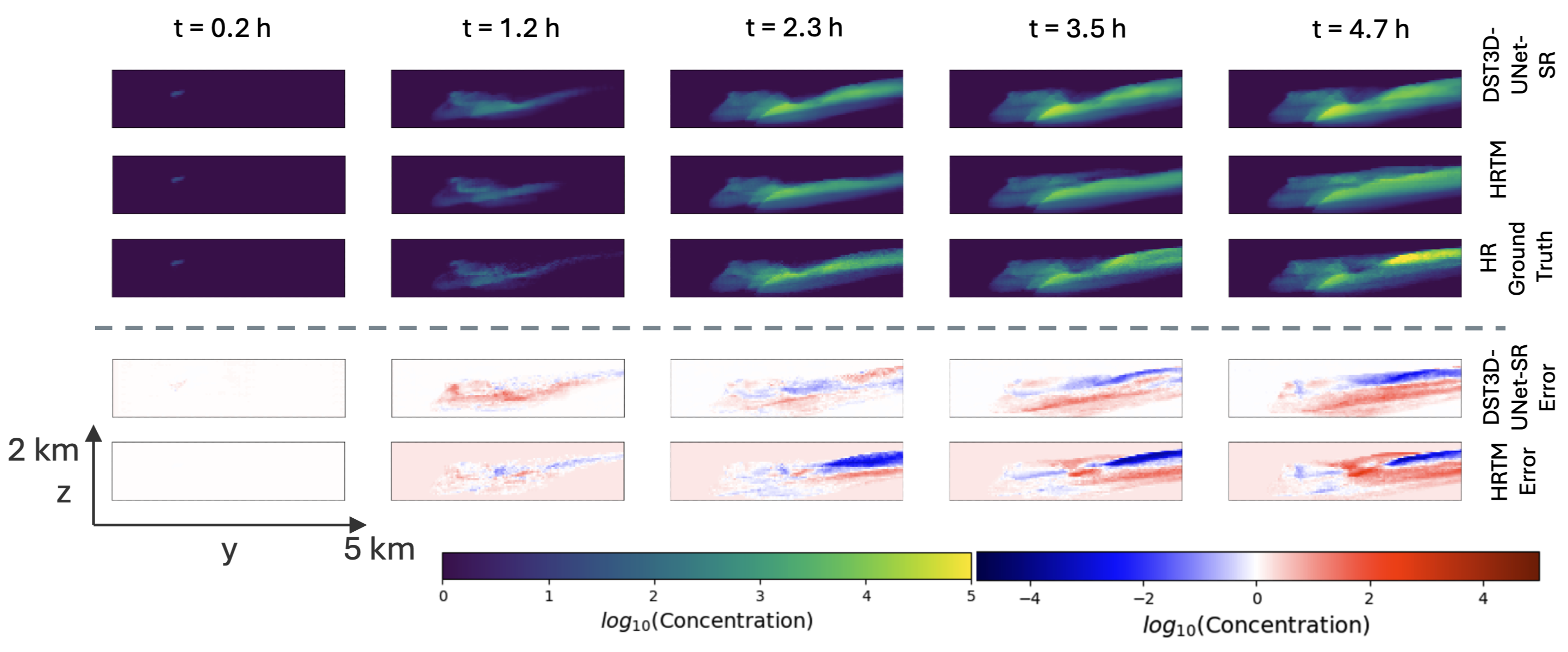} \caption{Averaged concentration in the \textit{y-z} plane at various time steps, comparing DST3D-UNet-SR and HRTM predictions (top two rows) with ground truth (third row). The errors are also shown in the figure (two bottom rows). The DST3D-UNet-SR error is calculated as the difference between DST3D-UNet-SR predictions and ground truth. Similarly, the HRTM error is calculated as the difference between HRTM predictions and ground truth. Variations are shown across the \textit{y} and \textit{z} dimensions within a range of 0 to 5 km and the \textit{z-}dimension ranging from 0 to 2 km.} \label{y_mean} 
        \end{figure*} 
        
        In Figures \ref{x_mean} to \ref{z_mean} we can see that qualitatively both, DST3D-UNet-SR and HRTM, perform well. The associated errors, presented as the difference between the predictions and ground truth give more insight. In high-resolution models like the HRTM, errors in early timesteps can rapidly propagate and grow because the model tries to predict both temporal evolution and fine spatial details simultaneously. This can result in exaggerated or over-predicted regions, particularly outside the main plume, as we see in the HRTM error row of figures \ref{x_mean},  \ref{y_mean} and \ref{z_mean}. The DST3D-UNet-SR’s approach of handling temporal and spatial resolution separately, starting with coarse temporal predictions and later refining them, allows it to avoid the common issues of noise and overemphasis on low-concentration regions. 

\begin{figure}[ht!] 
        \centering \noindent\includegraphics[width=0.5\textwidth]{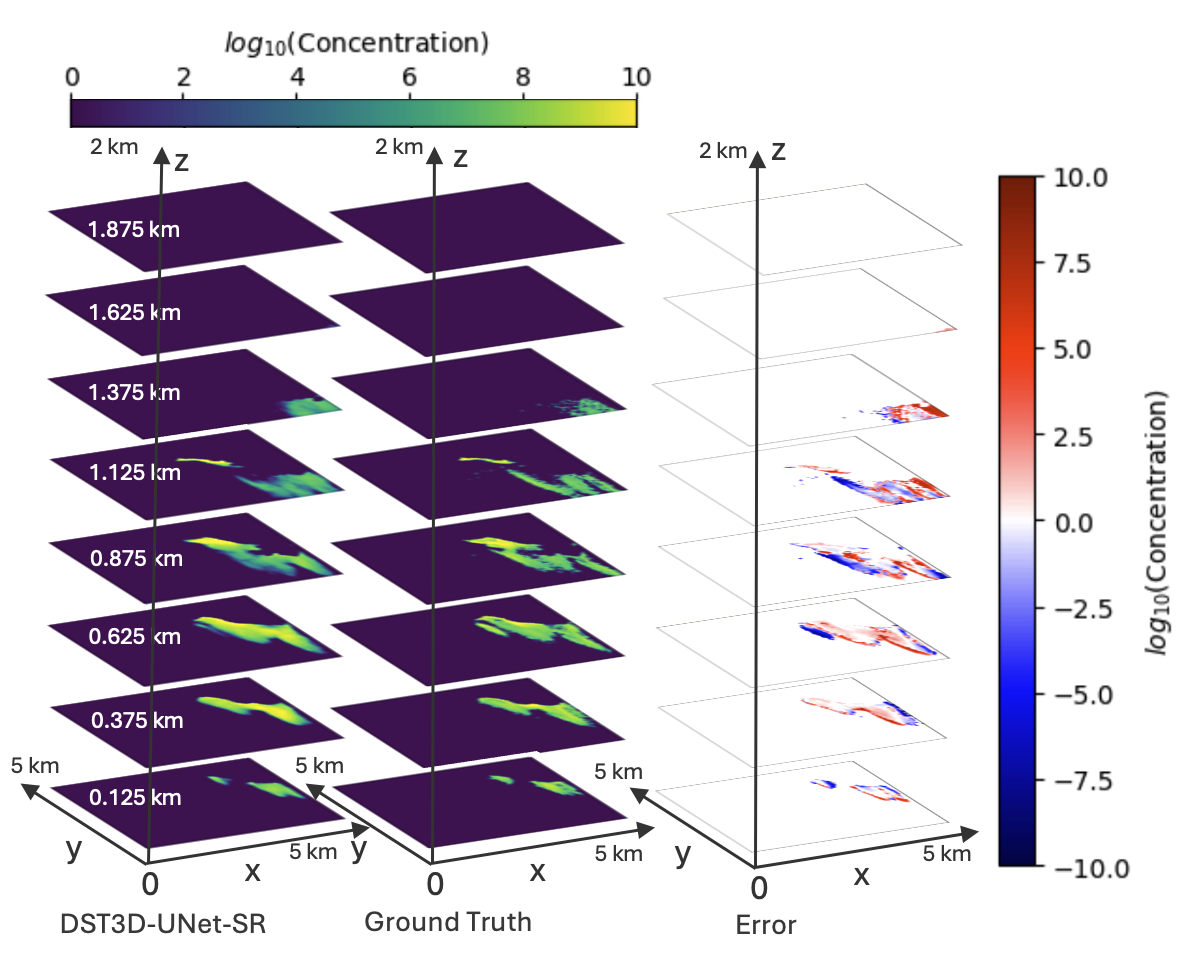} \caption{Slices along the \textit{z-}direction, 120 minutes post-gas release, illustrating DST3D-UNet-SR predictions (left) versus ground truth (middle). The error shown in the figure (right) is calculated as the difference between DST3D-UNet-SR predictions and ground truth. Variations are shown across the \textit{x} and \textit{y} dimensions within a range of 0 to 5 km and the \textit{z-}dimension ranging from 0 to 2 km. The spatial resolution is 0.04 km for the \textit{x-} and \textit{y-}dimensions and 0.02 km for the \textit{z-}dimension. Concentrations are recorded at the cell center. In the \textit{z-}direction, a total vertical extent of 2 km is represented with slice intervals of 0.125 km.} \label{z_slices} 
        \end{figure}

        Additionally, Figure \ref{z_slices} presents slices along the \textit{z-}direction, captured 120 minutes after the gas release. These slices compare the DST3D-UNet-SR predictions (left column) with the ground truth (middle column). The right column displays the error, calculated as the difference between the predictions and the ground truth. This figure demonstrates variations across the \textit{x-} and \textit{y-}dimensions. In the \textit{z-}direction, the vertical extent of 2 km is represented with slice intervals of 0.125 km.

		\section{Performance Metrics} \label{sec:performance}

        To quantify the performance of the models, we present four metrics: mean squared error (MSE), intersection over union (IoU), structural similarity index measure (SSIM), and conservation of mass (CM). Each of these metrics provides insight into different aspects of the model performance. While MSE and CM focus on the accuracy of the predicted values, IoU and SSIM assess the model's ability to accurately identify the average plume location and outline. All metrics are computed across the entire volume and over time.
        
        MSE measures the average squared difference between predicted and true values, indicating overall accuracy \cite{goodfellow2016deep}. IoU measures the overlap between predicted and ground truth binary masks; in our data, a reasonable threshold that preserved the plume outline was $\log_{10}(\text{Concentration}) = 1$, reflecting the model's ability to identify the correct plume location \cite{jaccard1912distribution}. SSIM assesses the similarity between predicted and true images based on luminance, contrast, and structure, providing a measure of perceptual quality \cite{wang2004image}. CM calculates the difference in total mass between predictions and ground truth, ensuring mass conservation in the model \cite{leveque2002finite}.

\begin{figure}[ht!]
        \centering \noindent\includegraphics[width=0.35\textwidth]{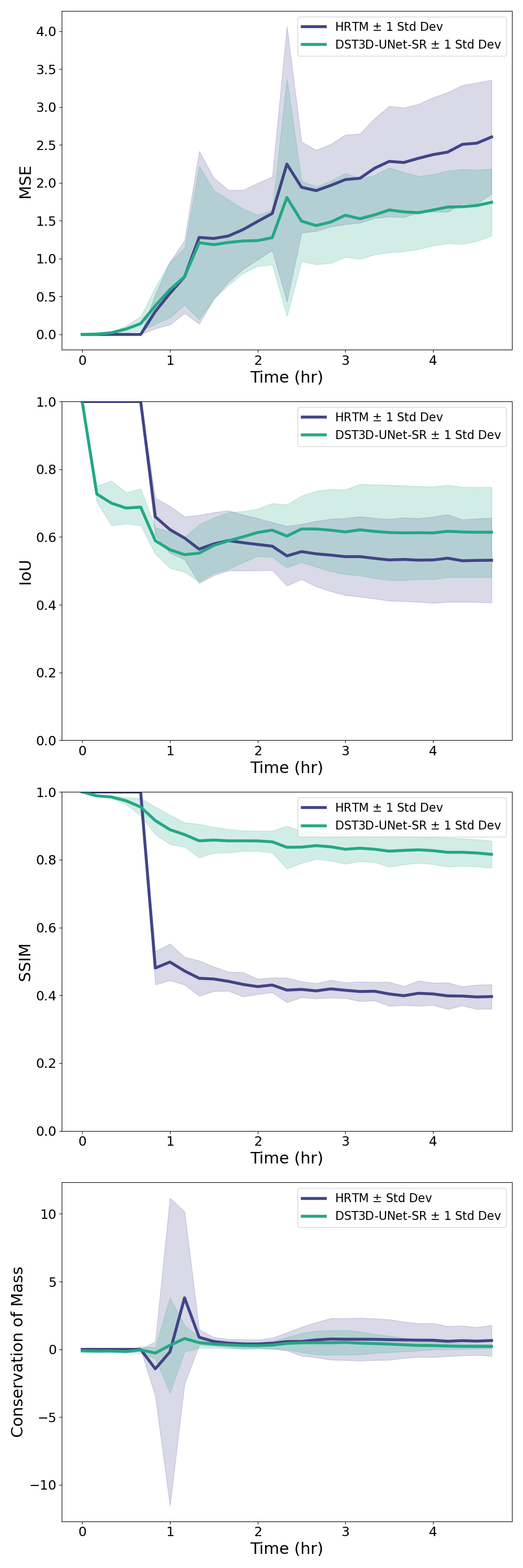} \caption{Performance metrics over time for DST3D-UNet-SR and HRTM. The metrics include MSE, IoU, SSIM, and CM. The continuous line represents the mean performance across 10 test simulations, while the shaded regions denote $\pm$ 1 standard deviation from the mean.} \label{metrics} 
        \end{figure}

        Figure~\ref{metrics} illustrates these performance metrics over time for both the DST3D-UNet-SR and the HRTM. 

        The performance metrics averaged over 33 time steps across 10 test runs are summarized in Table \ref{metrics_table}. 

        \begin{table}[ht]
            \centering
            \scriptsize
            \caption{Performance metrics averaged over 33 time steps across 10 test runs. The arrow direction indicates whether higher or lower metric values represent better performance. Note that the HRTM performance is overall inferior, despite the first 5 time steps being high-resolution ground truth for this model.}
            \begin{tabular}{lcccc}
                \toprule
                \textbf{Model/Metric} & \textbf{MSE $\downarrow$} & \textbf{IoU} $\uparrow$ & \textbf{SSIM} $\uparrow$ & \textbf{CM} $\downarrow$ \\
                \midrule
                \textbf{DST3D-UNet-SR} & \textbf{1.2 $\pm$ 0.4} & \textbf{0.62 $\pm$ 0.08} & \textbf{0.84 $\pm$ 0.03} & \textbf{0.3 $\pm$ 0.3} \\
                \textbf{HRTM} & 1.5 $\pm$ 0.4  & 0.62 $\pm$ 0.09 & 0.51 $\pm$ 0.02 & 0.6 $\pm$ 0.8 \\
                \bottomrule
            \end{tabular} \label{metrics_table}
        \end{table}
 
        Figure \ref{metrics} provides a detailed view of how these metrics evolve over time showing that the DST3D-UNet-SR model outperforms the HRTM overall, except during the initial five time steps when the HRTM benefits from high-resolution ground truth data. However, once the models rely on predictions alone, the DST3D-UNet-SR demonstrates superior performance across multiple metrics.

		The DST3D-UNet-SR model effectively captures the 3D nature of plume dispersion through its convolutional operations, while the LSTM layers accurately learn and predict the temporal evolution of the plume for up to five hours. The separation of temporal and spatial refinement helps avoid the error amplification that is more prominent in the HRTM model. This advantage is particularly evident in the consistently lower MSE values and higher SSIM scores of the DST3D-UNet-SR compared to the HRTM.

        The MSE plot reveals two distinct slopes: an initial sharp rise as both models transition from relying on ground truth to making predictions, followed by a more gradual slope where the DST3D-UNet-SR’s errors grow slower than the HRTM’s, indicating better control over error propagation. Despite similar spikes in MSE around the 2-hour mark, the DST3D-UNet-SR maintains a lower overall error due to better generalization across time steps.

        In the initial stages of dispersion, the concentration gradients are larger, making it more challenging to capture the plume’s exact location. In this phase, even small shifts in the predicted plume’s position can lead to a significant reduction in overlap, causing the IoU to drop sharply. Nevertheless, the IoU scores indicate that both models perform similarly in terms of spatial overlap with the ground truth, although DST3D-UNet-SR maintains slightly better performance over time.
        
        For the SSIM metric both models show a similar abrupt decline as they struggle to maintain the correct spatial patterns and texture of the plume during the initial transition. Rapid changes in concentration gradients and plume morphology can lead to distortions in the predicted structure, reducing the SSIM score. However, the decline for the DST3D-UNet-SR is less severe. Both models’ SSIM values plateau after one hour, but the DST3D-UNet-SR significantly outperforms the HRTM.
        
        With respect to CM, the DST3D-UNet-SR again shows superior and more robust performance in maintaining the physical properties of the plume over extended periods. The spikes in CM seen in both models around the 1-hour mark highlight the models’ responses to dynamic plume behavior. While the DST3D-UNet-SR also experiences a spike, it is less pronounced than that of the HRTM. This is because the SRM controls the spread of the plume by focusing on enhancing spatial details in regions where the temporal behavior is already well-predicted. The dual-stage process allows the model to mitigate large errors that could arise from incorrect predictions in peripheral regions.
        
        In terms of computational efficiency, the DST3D-UNet-SR model demonstrates a significant advantage over the HRTM, achieving single-time-step computation times nearly five times faster (0.35 ± 0.01 seconds for DST3D-UNet-SR model compared to 1.58 ± 0.01 seconds for HRTM). This efficiency corresponds to a speedup of three orders of magnitude relative to the original simulation runtime (2.25 hours), compared to the two orders of magnitude improvement achieved by the HRTM. These attributes make DST3D-UNet-SR model an excellent choice for rapid response scenarios or iterative processes requiring a large number of simulations, where both speed and accuracy are critical.

        \section{Validation Against Simulation Sensor Data} \label{validation}

        To further assess the predictive accuracy of the DST3D-UNet-SR and HRTM models, we compared their outputs to the sensor data generated by Aeolus simulations described in Section \ref{Data}. Although direct comparison with experiments is not possible due to the sparsity of field data, the agreement of simulation conditions with the REACT experiment provides a robust reference point for evaluating model accuracy.
        
\begin{figure}[htbp] \centering \noindent\includegraphics[width=0.5\textwidth]{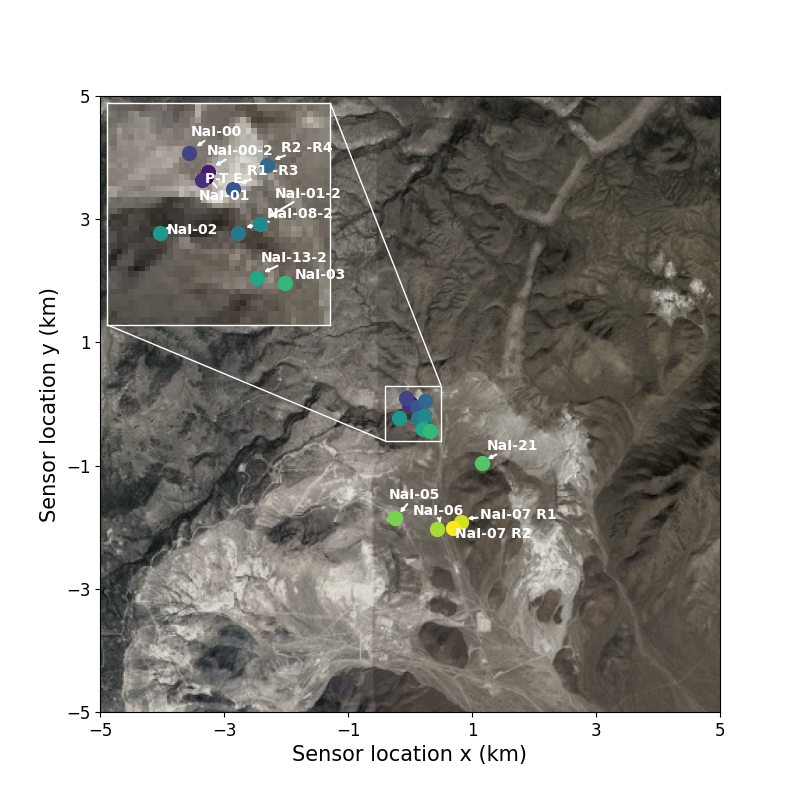} \caption{Real-time xenon sensors were deployed to monitor radiotracer releases at the Nevada National Security Site (NNSS) during the REACT (RElease ACTivity) experiment in October 2022.} \label{sensor_location}
        \end{figure}

        In these evaluations, sensor locations within the Aeolus simulation grid were identified and treated as ground truth references (see Figure \ref{sensor_location}). Since sensor height was not accounted for, concentration predictions were averaged across the vertical ($z$) dimension. To evaluate the models' predictive accuracy at the sensor sites, concentration values were compared at discrete locations over time rather than across the entire field, enabling a focused analysis of model performance under conditions replicating actual sensor placements during the experiment. Notably, as the models were originally trained on Aeolus simulations, no additional training or new simulations were required for this study.
        
\begin{figure*}[ht!] 
        \centering \noindent\includegraphics[width=\textwidth]{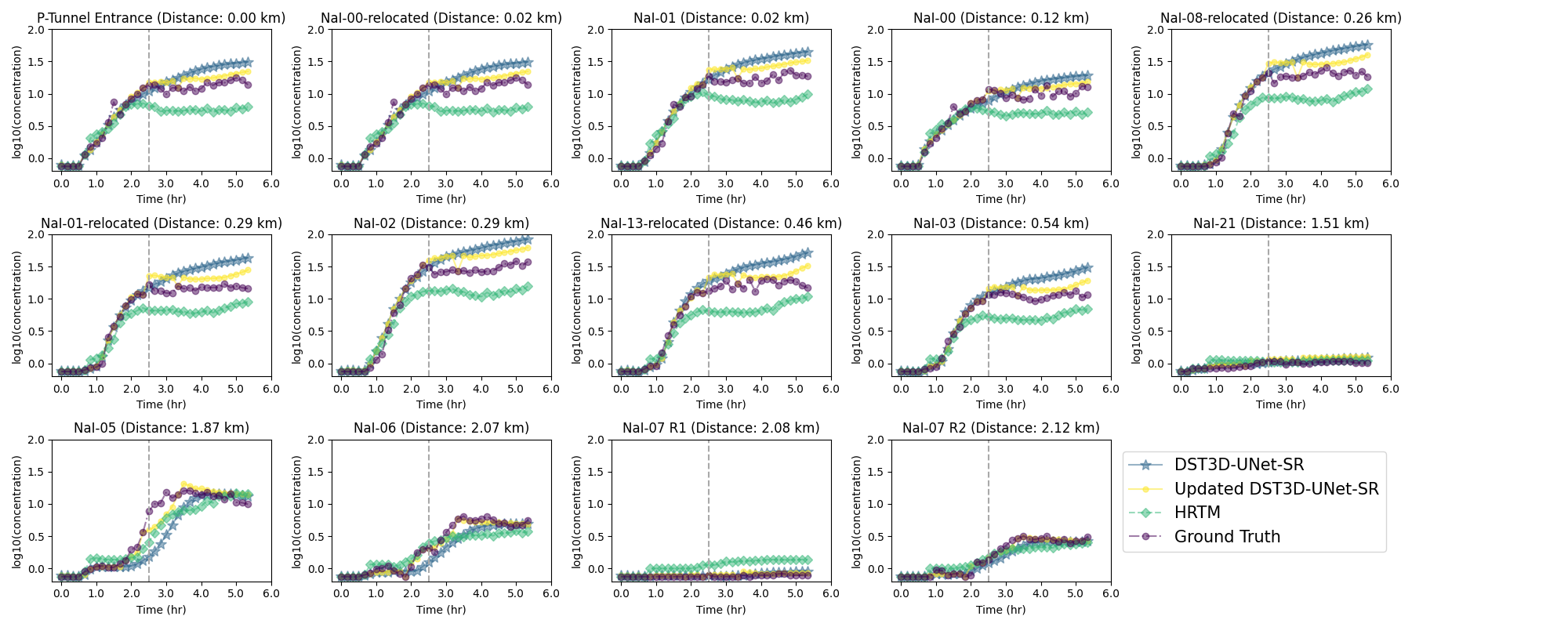} \caption{Validation of DST3D-UNet-SR and HRTM performance against Aeolus simulations (Ground Truth). The plot shows the z-averaged log10(concentration) detected by each sensor as a function of time, reported as the mean across the 10 test runs. The vertical grey dashed line qualitatively marks the separation between areas of good and poorer performance, approximately 2.5 hours after the release. Updated DST3D-UNet-SR refers to the predictions generated by DST3D-UNet-SR when it is provided with three additional data points as inputs at specific time intervals.} \label{sensor_detection} 
        \end{figure*}

        Figure \ref{sensor_detection} shows the z-averaged log10(concentration) detected by each sensor over time, reported as the mean across the 10 test runs. The gray dashed line, approximately 2.5 hours after the release, acts as a qualitative threshold indicating the transition from reliable to less reliable model predictions. The figure includes ground truth values based on Aeolus simulations (purple), predictions from DST3D-UNet-SR (blue), and predictions from HRTM (green). Additionally, the updated DST3D-UNet-SR (yellow) represents predictions generated by DST3D-UNet-SR when provided with three additional data points as inputs at specific time intervals (1h, 1.5h, and 2.5h). These updates reflect a typical operational scenario in which initial predictions requiring rapid responses are refined as new observational data become available.

        For sensors closest to the source (0 km to ~0.54 km), both models initially agree closely with the ground truth during the first two hours. Between 2 and 2.5 hours, the DST3D-UNet-SR remains more accurate, closely following the ground truth, while the HRTM starts to under-predict. Beyond this point, DST3D-UNet-SR begins to slightly over-predict concentrations relative to the ground truth, while the HRTM continues to under-predict. This discrepancy is likely due to sensor concentrations being recorded at discrete spatial points, causing the HRTM to generally under-predict concentrations inside the plume while over-predicting concentrations outside of it, as discussed in Section \ref{results}. However, while the HRTM captures the concentration plateau reached after 2.5 hours, DST3D-UNet-SR predictions continue to increase smoothly over time. 

        The updated DST3D-UNet-SR simulations, incorporating additional data points, show a smaller gap between predictions and the ground truth. The updates also enable the model to capture the pseudo-steady-state concentration plateau reached after 2.5 hours, as shown by the Aeolus simulations. While these updates improve predictions, particularly in high-concentration regions near the source, slight over-predictions persist. These updated predictions are included to illustrate the potential for improving model performance with new data. However, a detailed study into updating methods and their broader impacts is beyond the scope of this paper and will be addressed in future work.
        
        For sensors located farther away (1.51 km to ~2.12 km), both models exhibit comparable performance, effectively capturing late-time behavior with greater accuracy. This improved performance at larger distances is likely attributed to the plume becoming more diffuse, resulting in smoother concentration gradients that are easier to predict. Additionally, averaging over larger areas at these distances diminishes the impact of smaller-scale inaccuracies, enabling both models to align more closely with the ground truth. For these distant sensors, the updated DST3D-UNet-SR predictions demonstrate enhancements relative to the original DST3D-UNet-SR and HRTM, though the improvement is less pronounced compared to sensors closer to the source.
        
		\section{Conclusion}
		
        The DST3D-UNet-SR model outperforms the baseline HRTM across all key metrics, including lower mean squared error (MSE) and conservation of mass (CM) deviations, alongside superior structural similarity index measure (SSIM) and intersection over union (IoU) scores. These results highlight its capacity to accurately replicate the spatiotemporal dynamics of atmospheric plume dispersion while adhering to fundamental physical conservation laws.
        
        A significant advancement of the DST3D-UNet-SR is its computational efficiency, achieving a fivefold reduction in single-time-step computation time compared to HRTM and a 45\% reduction in training time (5.5 to 3 hours). This balance between computational speed and predictive fidelity underscores its suitability for applications requiring rapid response, such as real-time plume dispersion modeling, iterative optimization, and uncertainty quantification.

        We demonstrated that the model can dynamically incorporate observational updates during inference, enhancing predictive accuracy at later stages, particularly near the source where concentrations are highest. This capability highlights the potential of DST3D-UNet-SR to adapt to evolving conditions in real-world applications. Future work will explore the scalability and effectiveness of this adaptive approach across diverse scenarios.

        Future efforts will explore inferring initial low-resolution concentration fields directly from scalar wind conditions, reducing reliance on pre-computed simulation data and enhancing operational flexibility. Moreover, future efforts will also focus on integrating topographic information to improve generalization across varying terrains.

        \section*{Acknowledgments}
        This work was performed under the auspices of the U.S. Department of Energy by Lawrence Livermore National Laboratory under Contract DE-AC52-07NA27344. Institutional release number LLNL-JRNL-2001564. This Low Yield Nuclear Monitoring (LYNM) research was funded by the National Nuclear Security Administration, Defense Nuclear Nonproliferation Research and Development (NNSA DNN R\&D). The authors acknowledge important interdisciplinary collaboration with scientists and engineers from LANL, LLNL, NNSS, PNNL, and SNL.
        
		\section*{Authors Declarations} \label{declarations}
        \subsection*{Conflict of Interest}
        The authors have no conflicts to disclose.
        \subsection*{Author Contributions} \label{contributions}
        \textbf{M. Giselle Fern\'andez-Godino}: Conceptualization (lead); Data curation (lead); Investigation (lead); Methodology (lead); Project administration (lead); Software (lead); Supervision (lead); Visualization (lead); Writing – original draft (lead); Writing – review \& editing (lead). \textbf{Wai Tong Chung}: Conceptualization (supporting); Supervision (supporting); Data curation (supporting); Methodology (supporting); Review \& editing (supporting). \textbf{Akshay Gowardhan}: Physics model subject matter expert. \textbf{Matthias Ihme}: Review \& editing (supporting). \textbf{Qingkai Kong}: Conceptualization (supporting), Supervision (supporting); Review \& editing (supporting). \textbf{Donald D. Lucas}: Conceptualization (supporting); Supervision (supporting); Investigation (supporting) Writing – review \& editing (supporting). \textbf{Stephen C. Myers}: Supervision (supporting); Review \& editing (supporting); Funding acquisition; Project administration; Resources.
        \subsection*{Declaration of generative AI and AI-assisted technologies in the writing process}
        During the preparation of this work the author(s) used Lawrence Livermore National Laboratory's LivChat based on OpenAI's GPT-4 in order to ensure that the writing meets academic standards and is free of errors. After using this tool/service, the author(s) reviewed and edited the content as needed and take(s) full responsibility for the content of the publication.
        \section*{Data Availability}
        The data that support the findings of this study are available from the corresponding author upon reasonable request.

		\bibliographystyle{IEEEtran} 
		\bibliography{bibliography}
		
		\appendix
		
		\subsection{DualStage Temporal 3D UNet-SR Model Architecture} \label{A}
		
		The DualStage Temporal 3D UNet-SR architecture is designed to enhance the temporal resolution and spatial detail of atmospheric dispersion simulations. The model consists of two stages: a low-resolution-based temporal module (TM) and a high-resolution-based spatial refinement module (SRM). Further details are included in subsections \ref{TM} and \ref{SRM}.

		\subsubsection{Temporal Module} \label{TM}
  
        The TM is constructed using a convolutional autoencoder architecture, incorporating ConvLSTM layers at its bottleneck to capture temporal dependencies within the input data. Its architecture comprises 3,214,401 trainable parameters. It operates on input data of shape $(\text{batch size}, 5, 8, 32, 32)$, representing a temporal window of five time steps, each with a spatial resolution of $8 \times 32 \times 32$.
		
		\textbf{Encoder:} 
		\begin{itemize} 
			\item \textbf{enc1:} \(\text{Conv3d}(5, 7 \times 16, 3, \text{padding}=1)\) followed by BatchNorm3d, ReLU
			activation, MaxPool3d(2), and Dropout3d(0.2). 
			\item \textbf{enc2:} \(\text{Conv3d}(7 \times 16, 7 \times 32, 3, \text{padding}=1)\) followed by BatchNorm3d,
			ReLU activation, MaxPool3d(2), and Dropout3d(0.2). 
			\item \textbf{enc3:} \(\text{Conv3d}(7 \times 32, 7 \times 64, 3, \text{padding}=1)\) followed by
			BatchNorm3d(2), ReLU activation, MaxPool3d(2), and Dropout3d(0.2). 
		\end{itemize}
		
		\textbf{ConvLSTM:} 
		\begin{itemize} 
			\item \textbf{convlstm:} \(\text{Conv3d}(7 \times 64, 7 \times 64, 3, \text{padding}=1)\). 
		\end{itemize}
		
		\textbf{Decoder:} 
		\begin{itemize} 
			\item \textbf{dec1:} \(\text{ConvTranspose3d}(7 \times 64, 7 \times 32, 2, \text{stride}=2)\) followed by
			BatchNorm3d, ReLU activation, and Dropout3d(0.2). 
			\item \textbf{dec2:} \(\text{ConvTranspose3d}(7 \times 32, 7 \times 16, 2, \text{stride}=2)\) followed by
			BatchNorm3d, ReLU activation, and Dropout3d(0.2).
			\item \textbf{dec3:} \(\text{ConvTranspose3d}(7 \times 16, 1, 2, \text{stride}=2)\) followed by ReLU
			activation. 
		\end{itemize}
		
		The TM generates single time-step predictions with an output shape of $(\text{batch size}, 1, 8, 32, 32)$, which are recursively fed back into the model in a sliding window framework consisting of five time steps. 
		
		\subsubsection{Spatial Refinement Module}  \label{SRM}
		
		The SRM processes the TM output, reshaping it to $(\text{batch size}, 8, 32, 32)$ and produces high-resolution output of shape $(\text{batch size}, 32, 128, 128)$. Its architecture has 951,873 trainable parameters.
		
		\textbf{Encoder:} 
		\begin{itemize} 
			\item \textbf{enc1:} \(\text{Conv3d}(1, 7 \times 16, 3, \text{padding}=1)\) followed by BatchNorm3d, LeakyReLU
			activation, and MaxPool3d. 
			\item \textbf{enc2:} \(\text{Conv3d}(7 \times 16, 7 \times 32, 3, \text{padding}=1)\) followed by BatchNorm3d, LeakyReLU
			activation, and MaxPool3d. 
		\end{itemize}
		
		\textbf{Adjust Channels for Skip Connections:} 
		\begin{itemize} 
			\item \textbf{adjust\_channels:} \(\text{Conv3d}(7 \times 16, 7 \times 32, 1)\)
			followed by BatchNorm3d and LeakyReLU activation. 
		\end{itemize}
		
		\textbf{Decoder:} 
		\begin{itemize} 
			\item \textbf{dec1:} \(\text{ConvTranspose3d}(7 \times 32, 7 \times 32, 3, \text{stride}=2, \text{padding}=1,
			\text{output\_padding}=1)\) followed by BatchNorm3d and LeakyReLU activation. 
			\item \textbf{dec2:} \(\text{ConvTranspose3d}(7 \times 32, 7 \times 16,
			3, \text{stride}=2, \text{padding}=1, \text{output\_padding}=1)\) followed by BatchNorm3d and LeakyReLU activation. 
			\item \textbf{dec3:}
			\(\text{ConvTranspose3d}(7 \times 16, 7 \times 8, 3, \text{stride}=2, \text{padding}=1, \text{output\_padding}=(1, 1, 1))\) followed by BatchNorm3d
			and LeakyReLU activation. 
			\item \textbf{dec4:} \(\text{ConvTranspose3d}(7 \times 8, 1, (1, 3, 3), \text{stride}=(1, 2, 2), \text{padding}=(0, 1, 1),
			\text{output\_padding}=(0, 1, 1))\) followed by LeakyReLU activation. 
		\end{itemize}
		
		\subsection{High-resolution Temporal Module Architecture} \label{B}
  
        The high-resolution temporal model (HRTM) employs a convolutional autoencoder architecture, similar to the TM, with ConvLSTM layers integrated at the bottleneck to effectively capture temporal dependencies within the input data. The HRTM is composed of 4,166,274 trainable parameters. The model operates on input data of shape $(\text{batch size}, 5, 32, 128, 128)$, representing a temporal window of five time steps, each with a spatial resolution of $32 \times 128 \times 128$.
		
		\textbf{Encoder:} 
		\begin{itemize} 
			\item \textbf{enc1:} \(\text{Conv3d}(5, \text{num\_layers} \times 32, 3, \text{padding}=1)\) followed by BatchNorm3d, ReLU activation, MaxPool3d, and Dropout3d.
			\item \textbf{enc2:} \(\text{Conv3d}(\text{num\_layers} \times 32, \text{num\_layers} \times 64, 3, \text{padding}=1)\) followed by BatchNorm3d, ReLU activation, MaxPool3d, and Dropout3d.
			\item \textbf{enc3:} \(\text{Conv3d}(\text{num\_layers} \times 64, \text{num\_layers} \times 128, 3, \text{padding}=1)\) followed by BatchNorm3d, ReLU activation, MaxPool3d, and Dropout3d.
		\end{itemize}
		
		\textbf{ConvLSTM:} 
		\begin{itemize} 
			\item \textbf{convlstm:} \(\text{Conv3d}(\text{num\_layers} \times 128, \text{num\_layers} \times 128, 3, \text{padding}=1)\).
		\end{itemize}
		
		\textbf{Decoder:} 
		\begin{itemize} 
			\item \textbf{dec1:} \(\text{ConvTranspose3d}(\text{num\_layers} \times 128, \text{num\_layers} \times 64, 2, \text{stride}=2)\) followed by BatchNorm3d, ReLU activation, and Dropout3d.
			\item \textbf{dec2:} \(\text{ConvTranspose3d}(\text{num\_layers} \times 64, \text{num\_layers} \times 32, 2, \text{stride}=2)\) followed by BatchNorm3d, ReLU activation, and Dropout3d.
			\item \textbf{dec3:} \(\text{ConvTranspose3d}(\text{num\_layers} \times 32, 1, 2, \text{stride}=2)\) followed by ReLU activation.
		\end{itemize}
		
		Like the TM, the HRTM produces single time-step predictions with an output shape of $(\text{batch size}, 1, 32, 128, 128)$, which are recursively fed back into the model in a sliding window framework consisting of five time steps. 

	\end{document}